\documentclass{article}
\pdfoutput=1
\usepackage{ijcai23}
\usepackage{times}
\usepackage{soul}
\usepackage{url}
\usepackage[hidelinks]{hyperref}
\usepackage[utf8]{inputenc}
\usepackage[small]{caption}
\usepackage{graphicx}
\usepackage{amsmath}
\usepackage{amsthm}
\usepackage{amssymb}
\usepackage{booktabs}
\usepackage{algorithm}
\usepackage{algorithmic}
\usepackage[switch]{lineno}

\usepackage{multirow}
\newcommand{\ie}{\textit{i}.\textit{e}.}
\newcommand{\eg}{\textit{e}.\textit{g}.}

\usepackage{color}
\usepackage{xcolor}
\usepackage{colortbl}
\definecolor{ours}{gray}{.95}

\title{Discrepancy-Guided Reconstruction Learning for Image Forgery Detection}
\author{
Zenan Shi$^{1,2}$\and Haipeng Chen$^{1,2}$\and Long Chen$^3$\and Dong Zhang$^{3,}$\thanks{Corresponding author: Dong Zhang.}
\affiliations
$^1$College of Computer Science and Technology, Jilin University \\
$^2$Key Laboratory of Symbolic Computation and Knowledge Engineering of Ministry of Education, \\ Jilin University\\
$^3$Department of CSE, The Hong Kong University of Science and Technology
\emails
\{shizn,~chenhp\}@jlu.edu.cn, \{longchen,~dongz\}@ust.hk
}
\begin{document}
\maketitle
\begin{abstract}
In this paper, we propose a novel image forgery detection paradigm for boosting the model learning capacity on both forgery-sensitive and genuine compact visual patterns. Compared to the existing methods that only focus on the discrepant-specific patterns (\eg, noises, textures, and frequencies), our method has a greater generalization. 
Specifically, we first propose a Discrepancy-Guided Encoder (DisGE) to extract forgery-sensitive visual patterns. DisGE consists of two branches, where the mainstream backbone branch is used to extract general semantic features, and the accessorial discrepant external attention branch is used to extract explicit forgery cues. 
Besides, a Double-Head Reconstruction (DouHR) module is proposed to enhance genuine compact visual patterns in different granular spaces. 
Under DouHR, we further introduce a Discrepancy-Aggregation Detector (DisAD) to aggregate these genuine compact visual patterns, such that the forgery detection capability on unknown patterns can be improved. 
Extensive experimental results on four challenging datasets validate the effectiveness of our proposed method against state-of-the-art competitors. 

\end{abstract}
\section{Introduction}
\label{intro}
\begin{figure*}[th!]
\centering
\includegraphics[width=.99\linewidth]{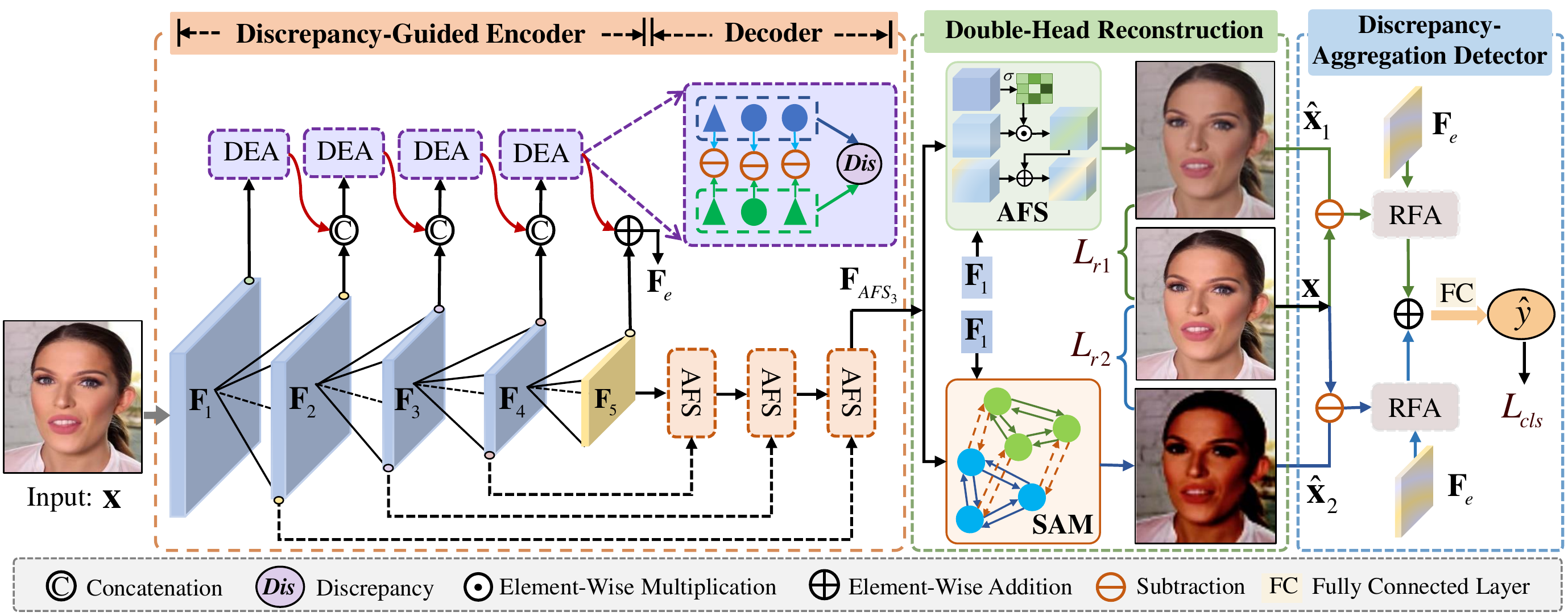}
\vspace{-2mm}
\caption{Overall architecture of our DisGRL, which mainly consists of: a Discrepancy-Guided Encoder (DisGE), a decoder, a Double-Head Reconstruction (DouHR) module, and a Discrepancy-Aggregation Detector (DisAD) head network for image forgery classification.}
\label{figure1}
\vspace{-3mm}
\end{figure*}
Advanced image co-editing and synthesis methods make it cushy for people to tamper with images~\cite{zhu2020domain,rombach2022high}. For example, objects and external properties of these objects in a given image can be completely interpolated via a few texts~\cite{kawar2022imagic}. Although these progressive methods can increase the diversity and interest of images, on the other hand, they cause a new problem that people's confidence in the information expressed in images is reduced~\cite{Recce,fei2022learning}. Besides, tampered images may also be used in some malicious occasions (\eg, fake news and deliberate slanders), thus bringing potential social harms~\cite{hu2021dynamic,zhang2022deep,sun2022dual,li2020face}. Therefore, exploring effective image forgery detection methods is urgent.

In recent years, thanks to the immense progress of image processing technologies based on the deep learning mechanism~\cite{he2016deep,zhang2020feature}, ample semantic features greatly improve the recognition accuracy of image forgery detection~\cite{wang2021representative,zhuang2022uia} on both the image-level and the pixel-level~\cite{jiang2020deeperforensics,sun2021domain,zhao2021multi}. However, off-the-shelf deep learning methods cannot achieve satisfactory results in the face of some challenging forgery cases (\eg, unusual tampering areas and marginal tampering clues). To address this problem and improve the accuracy, some recent approaches use specific operators (\eg, BayerConv~\cite{chen2021image}, Sobel operator~\cite{zenanshi2022pretraining}, and frequency filter~\cite{fei2022learning}) to extract discrepant-specific patterns (\eg, noises, textures, and frequencies) as a supplementary for the recognition model. What these methods have commonly is that they use a mainstream backbone network and an additional auxiliary network to extract implicit and explicit semantic features, respectively~\cite{sun2022dual,wang2021representative,zhuang2022uia,zenanshi2022pretraining,chen2021image,zhao2021multi}. However, these methods usually need to generate temporary supervisions via intermediate feature maps, which are inherently not accurate enough, thus hurting the recognition effectiveness. 

What's more, the generalization capacity of existing methods is somewhat limited -- due to their overemphasis on the consequence of the explicit discrepant (\ie, the tampered region) features, which limits their latent usage scope.
To be specific, only learning some specific types of tampering patterns is far from pragmatism, because we cannot suppose tampering manners
~\cite{Recce,yoshihashi2019classification,chen2021local}. To improve the generalization capacity, it is helpful to learn a set of compact visual patterns, which inherently contain some general image properties, \eg, the concurrent local textures, the consistent regional resolutions, and the continuous bright changes~\cite{robert2018hybridnet,Recce,yoshihashi2019classification}. For image forgery detection, to achieve this goal, some work demonstrated that image reconstruction is an effective approach~\cite{wang2022uformer,li22cross}. The reconstructed output has rich compact patterns and suppresses local forgery regions. 
However, the existing methods are usually equipped with a single reconstruction head, which suffers from problems of tedious feature representations and inadequate reasoning ability.

In this paper, we propose a novel image forgery detection paradigm, named Discrepancy-Guided Reconstruction Learning (DisGRL), to improve the model learning capacity on both the forgery-sensitive and the {genuine compact visual patterns}. As illustrated in Figure~\ref{figure1}, DisGRL consists of four components: a Discrepancy-Guided Encoder (DisGE), a decoder, a Double-Head Reconstruction (DouHR) module, and a Discrepancy-Aggregation Detector (DisAD) head network. 

Specifically, the proposed DisGE (\textit{ref.}~Sec.~\ref{sec3:1}) is used to extract {forgery-sensitive visual patterns}, which consists of two branches: a mainstream backbone branch is used to extract the general semantic features, and an accessorial discrepant external attention branch is used to extract the explicit forgery visual cues. 
Thereby, DisGE is more suitable for the image forgery detection task than the common backbone networks (\eg, convolutional neural networks and vision transformer).
In the decoder, three progressive attention feature selection modules are employed in ${\bf{F}}_i$ to connect feature maps from the corresponding encoder network layer, which finally has the same scale as ${{\bf{F}}_2}$.
We also propose a DouHR module (\textit{ref.}~Sec.~\ref{sec3:2}) based on the decoder, which can enhance the genuine compact visual patterns in two separate granular spaces via an image reconstruction manner. To be specific, in the DouHR module, an Attention-guidance Feature Selection (AFS) procedure and a Similarity Aggregation Module (SAM) are used to extract the {vision-based} and the {reasoning-based} genuine compact visual patterns, respectively. 
Based on the DouHR, we further introduce a DisAD head network (\textit{ref.}~Sec.~\ref{sec3:3}) for image forgery classification, which can aggregate the obtained genuine compact visual patterns via a Reconstruction-guidance Feature Aggregation (RFA) module, resulting in an improved forgery detection capability on unknown patterns. 
Therefore, compared to the existing methods that only focus on these discrepant-specific patterns, our proposed DisGRL has a stronger generalization capability. To demonstrate the superiority of DisGRL, extensive experiments are carried out on four commonly used yet challenging face forgery detection datasets. Results validate that our DisGRL can achieve state-of-the-art performance on both seen and unseen forgeries.

Our contributions are as follows: 1) We propose a novel DisGRL for image forgery detection, which contains three proposed components for learning both forgery-sensitive and genuine compact visual patterns. 2) Extensive experimental results on four challenging datasets validate that DisGRL can achieve state-of-the-art performance against competitors.
\section{Related Work}
Image forgery can be viewed as a game of AI \emph{v.s.} AI since the majority of detection technologies are based on deep learning. In the past, many efforts have been made to improve the performance of natural/face image forgery detection~\cite{fei2022learning,haliassos2021lips,zhang2022unabridged,gu2022region,zhang2021detecting}. Extensive work uses a two-branch architecture to mine specific forgery patterns, such as noises or frequency domain features in combination with RGB spatial data, in light of the fact that altered images are getting more visually realistic~\cite{li2022wavelet,chen2021local,masi2020two,qian2020thinking,li2021frequency,wang2022objectformer}.
SOLA~\cite{jia2022exploring} fuses multimodal features from RGB and high-frequency features extracted by a DCT transformation in an extra branch for more general representations.
As a complementary of RGB, the model in~\cite{fei2022learning} introduces subtle noise features via learnable high pass filters with anomalies in local regions also performed well in unseen forgeries~\cite{zhang2020feature,yan2023progressive,zhang2022deep}. Despite their remarkable performance, their models for obtaining specific forgery patterns only reflect certain aspects of the forgery, which might lead to model bias or sub-optimization.

Recently, some advanced methods are proposed to improve the model generalization capacity such as exploiting contrastive learning to guide the recognition model focus on local content inconsistencies~\cite{sun2022dual,shi2023transformer,zhang2022graph}, introducing domain adaptation to alleviate overfitting on a single domain~\cite{rao2022towards,sun2021domain,rao2021self}, and/or enhance feature representation with an information-theoretic self-information metric for forgery detection~\cite{sun2022information}. These methods achieve both great performances under intra-dataset (\ie, seen) and cross-domain (\ie, unseen) evaluations. Unlike these methods that explore the local level inconsistencies, our method focuses more on forgery-sensitive and genuine compact visual patterns, which can improve both model’s accuracy and generalization.
\section{Our Approach}
\label{sec3}
DisGRL is proposed to improve the model learning capacity on both the forgery-sensitive and the genuine compact visual patterns. Our contributions lie in presenting: a Discrepancy-Guided Encoder (DisGE), a Double-Head Reconstruction (DouHR) module, and a Discrepancy-Aggregation Detector (DisAD) head network for image forgery classification. An overview architecture of DisGRL is illustrated in Figure~\ref{figure1}. The input is an RGB image ${\bf{X}}$, and the output is a binary predicted label ${{{\hat y}}}$, which indicates whether the input image is forged or not.
In the following, we detail the implementations of each proposed component.
\subsection{Discrepancy-Guided Encoder (DisGE)} 
\label{sec3:1}
To capture forgery-sensitive visual patterns, we propose a DisGE, which consists of two parallel branches, where the mainstream backbone based on Xception network~\cite{chollet2017xception}  is used to extract multi-level semantic features, \ie, ${\bf{F}}_i~(i=1,2,\dots,5)$, and the \textbf{Discrepancy External Attention (DEA)} branch is applied to different level feature to extract explicit discrepant-specific pattern, which are usually subtle and occur in local regions. As shown in Figure~\ref{figure1}, features from different Xception layers are combined in a cascaded manner by DEA block. The specific operation of each DEA block's output ${{\bf{D}}_i}$ is expressed as:
\begin{equation}
\label{eq1}
{{\bf{D}}_i} = \left\{ \begin{array}{l}Dea({{\bf{F}}_i}),\; ~~~~~~~~~~~~~~~~~~~~~~~ i = 1 \\
Dea(Cat({{\bf{D}}_{i - 1}},{{\bf{F}}_i})),\quad i \in [2,3,4]
\end{array} \right.
\end{equation}
where $Dea(\cdot)$ and $Cat(\cdot)$ denote each DEA block and feature concatenation along the channel dimension, respectively. 

\begin{figure}[t]
\centering
\includegraphics[width=.99\linewidth]{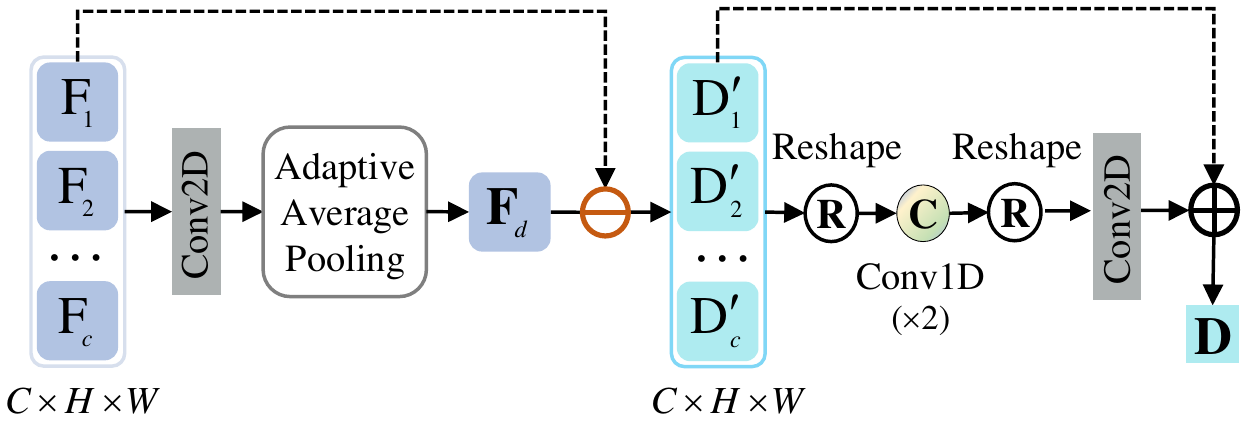}
\vspace{-2mm}
\caption{Illustration of the proposed Discrepancy External Attention (DEA) block, which is proposed to extract the forgery-sensitive visual patterns in the Discrepancy-Guided Encoder network.}
\label{figure2}
\vspace{-3mm}
\end{figure}

For each DEA block, as shown in Figure~\ref{figure2}, we first apply a $3 \times 3$ convolutional layer on the input feature maps ${\bf{F}} \in {\mathbb{R}^{C \times H \times W}}$ with the same channel size $C$.
Then, an adaptive average pooling is used to obtain the pooled features ${\bf{F}}_d$. 
After that, the differentiated maps can be obtained through ${\bf{D'}} = {\bf{F}} - {{\bf{F}}_d}$ to extract the discrepant information. Inspired by~\cite{guo2022beyond}, two 1D convolutions that share the same parameters are further introduced to characterize the global features of the entire map. Concretely, given a differentiated input $\bf{D'} \in {\mathbb{R}^{\textit{C} \times \textit{H} \times \textit{W}}}$, after reshaping and 1D convolution, feature maps are up-sampled four times in channel size. And 1D convolution and a reshape function are applied again to restore the original feature map size. Finally, the output feature map $\bf{D}$ can be obtained through a 1 $\times$ 1 convolution and a residual connection.
\subsection{Double-Head Reconstruction (DouHR)}
\label{sec3:2}
Reconstruction learning has been proven to be beneficial to several image forgery detection works by exploring rich compact visual patterns~\cite{Recce,wang2022uformer,li22cross}. 
In this work, we propose DouHR module based on the decoder to enhance the genuine compact visual patterns in two separate granular spaces (\ie, AFS and SAM) via an image reconstruction manner, such that the model can not only learn a rich genuine compact visual pattern but also further suppress the visual representation of the local forgery regions. 
As shown in Figure~\ref{figure1}, besides ${{\bf{\hat X}}_1}$ that is generated by an {\textbf{Attention-guidance Feature Selection (AFS)}} procedure in extracting the vision-based genuine compact visual patterns via convolutions, we introduce an extra {\textbf{Similarity Aggregation Module (SAM)}} for extracting the reasoning-based genuine compact visual patterns via secondary reconstruction ${{\bf{\hat X}}_2}$. The DouHR module can be formulated as:
\begin{equation}
\label{eq2}
{{\bf{\hat X}}_1} =  {\rm{AFS}}({{\bf{F}}_{AF{S_3}}},{{\bf{F}}_1}),{{\bf{\hat X}}_2} =  {\rm{SAM}}({{\bf{F}}_{AF{S_3}}},{{\bf{F}}_1}),
\end{equation}
where ${{\bf{F}}_{AF{S_3}}}$ indicates the output feature maps of the third {\textbf{AFS}} procedure in the decoder. In DouHR, we adjust the number of channels from the output of the SAM and AFS modules to 3 by applying a 1 × 1 convolution. After that, we use bilinear interpolation to adjust the feature map size to match the input image size.

\textbf{AFS}. In the decoder, three AFS modules receive the output of the previous AFS module and feature maps of the corresponding level in the mainstream backbone as input. For example, the inputs to the third AFS are ${{\bf{F}}_{AF{S_2}}}$ and ${{\bf{F}}_2}$. In DouHR, AFS receives the ${{\bf{F}}_{AF{S_3}}}$ and as ${{\bf{F}}_1}$ input. The concatenating operation $Cat( \cdot )$ is first carried out on ${{\bf{F}}_{AF{S_3}}}$ and ${{\bf{F}}_1}$ in the channel dimension, \ie, ${{\bf{\tilde F}}} = Cat({{\bf{F}}_{AF{S_3}}},{{\bf{F}}_1})$, followed by 
a depthwise separable convolution ${f_{d3}}$ to obtain attention map ${{\bf{A}}_{att}}$ with the same shape as input features and suppress the unimportant region of feature information transmitted by decoder
output, so that model pays more attention to the genuine compact visual patterns. Finally, a residual connection operation is applied to obtain the output. The above process can be expressed as follows:
\begin{equation}
\label{eq3}
{\begin{array}{c}
{{\bf{A}}_{att}} = \sigma ({f_{d3}}({\bf{\tilde F}})),\\
{\bf{A}} = {f_{d3}}({f_{d3}}({\bf{\tilde F}}) \odot {{\bf{A}}_{att}}) + {f_{c3}}({\bf{\tilde F}}),
\end{array}}
\end{equation}
where ${f_{c3}}$ and $\sigma(\cdot)$ are the 3 $\times$ 3 convolution layer and sigmoid activaion  function, respectively. Other AFS procedures are calculated in a similar way.

\begin{figure*}[t]
\centering
\includegraphics[width=0.80\linewidth]{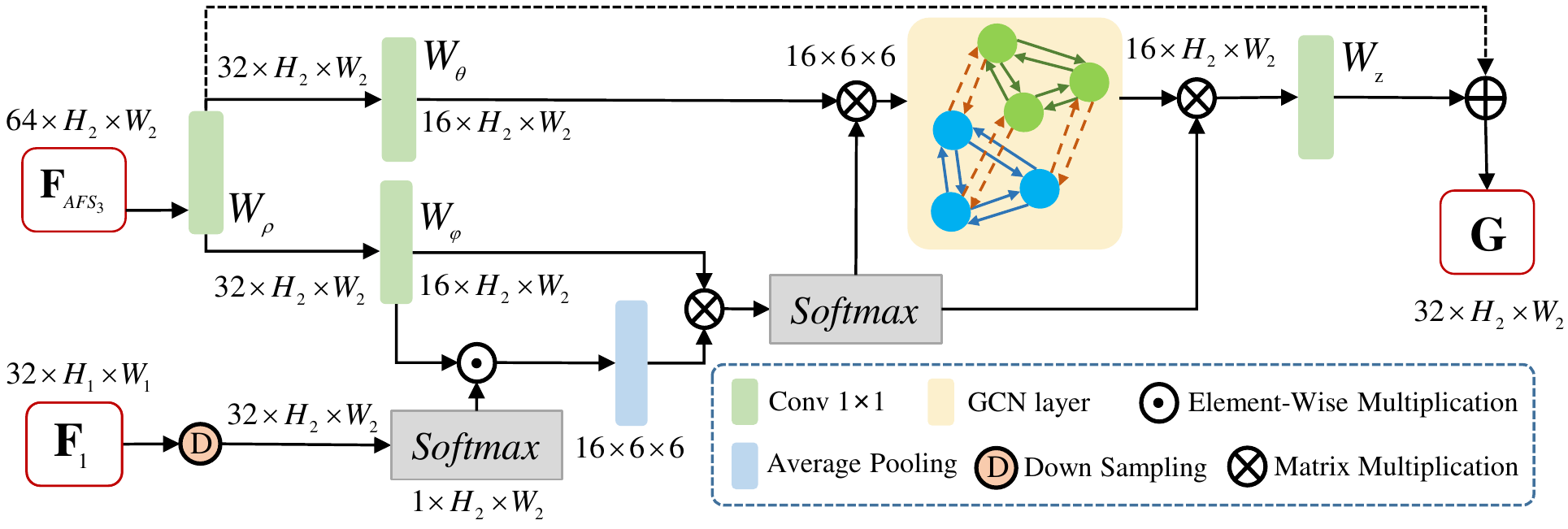}
\vspace{-2mm}
\caption{Illustration of Similarity Aggregation Module (SAM), which can extract the reasoning-based genuine compact visual patterns.}
\label{figure3}
\vspace{-3mm}
\end{figure*}
\textbf{SAM.}
To inject detailed global features into high-level semantic features using a global reasoning reconstruction, inspired by~\cite{dong2021polyp}, we introduce non-local operation under graph convolution operation~\cite{lu2019graph,zhang2022graph} to implement SAM. As shown in Figure~\ref{figure3}, for the given feature map ${{\bf{F}}_{AF{S_3}}}$, we first apply three 1 $\times$ 1 convolutions (\ie, ${\bf{W}_\rho}, \bf{W}_\theta$, and ${\bf{W}_\varphi}$) to reduce the channel dimension into 16, and obtain feature maps ${\bf{F}_\theta }$, ${\bf{F}_\varphi }$, which can be expressed as:
\begin{equation}
\label{eq4}
{\bf{F}_\theta } = {\bf{W}_\theta } ({\bf{W}_\rho }({{\bf{F}}_{AF{S_3}}})),~{\bf{F}_\varphi } = {\bf{W}_\varphi } ({\bf{W}_\rho }({{\bf{F}}_{AF{S_3}}})).
\end{equation}
For ${{\bf{F}}_1}$, we down-sample it to the same size as ${\bf{W}_\rho }({{\bf{F}}_{AF{S_3}}})$. Then we apply a Softmax function along the channel dimension and calculate the element-wise multiplication with  $ {\bf{F}_\varphi }$ for assigning different weights to different pixels and increasing the weight of edge pixels. And an adaptive pooling operation $Avp(\cdot)$ is utilized to reduce the displacement of features. In summary, the processing can be formulated as: 
\begin{equation}
\label{eq5}
{\bf{F}_{w}} = Avp({\bf{F}_\varphi } \odot {\mathop{\rm softmax}\nolimits} (D({{\bf{F}}_1}))), 
\end{equation}
where $D(\cdot)$ and $\rm{softmax}(\cdot)$ denote the down-sampling and Softmax functions, respectively.
After that, the matrix multiplication and Softmax function are used to establish the correlation between ${\bf{F}_\varphi }$ and ${\bf{F}_{w}}$, which can be expressed as:
\begin{equation}
\label{eq6}
{\bf{F}_{cor}} = {\mathop{\rm softmax}\nolimits} ({\bf{F}_{w}} \otimes {({\bf{F}_\varphi })^T}).
\end{equation}
The correlation attention map ${\bf{F}_{cor}}$ is multiplied with the feature map ${\bf{F}_\theta }$, and the resulting map is fed to the graph convolutional newtwork (GCN). Same to~\cite{dong2021polyp}, reconstructing the graph domain features into the original structural features as follows:
\begin{equation}
\label{eq7}
{\bf{G}'} = {\bf{F}_{cor}}^T \otimes GCN({\bf{F}_{cor}} \otimes {\bf{F}_\theta }).
\end{equation}
Finally, the reconstructed features $\bf{G}'$ are combined with the  features ${{\bf{W}_\rho }({{\bf{F}}_{AF{S_3}}})}$ to obtain the output ${\bf{G}}$:
\begin{equation}
\label{eq8}
    {\bf{G}} = {\bf{W}_\rho }({{\bf{F}}_{AF{S_3}}}) + {\bf{W}_z}({\bf{G}'}),
\end{equation}
where ${\bf{W}_z}$ denotes 1 $\times$ 1 convolution. 

\subsection{Discrepancy-Aggregation Detector (DisAD)}
\label{sec3:3}
The double-head reconstructed forged images essentially differ from the input forged images in visual appearance~\cite{Recce}. To further explore the probable forgery regions within reconstructed images, based on the DouHR module, we further introduce a DisAD head network via two \textbf{Reconstruction-guidance Feature Aggregation (RFA)} modules to aggregate the obtained genuine compact visual patterns (\ie, ${{\bf{\hat X}}_1}$ of AFS and ${{\bf{\hat X}}_2}$ of SAM), resulting in an improved forgery detection capability on unknown patterns (\ie, the greater generalization capability).

\begin{figure}[t]
\includegraphics[width=.99\linewidth]{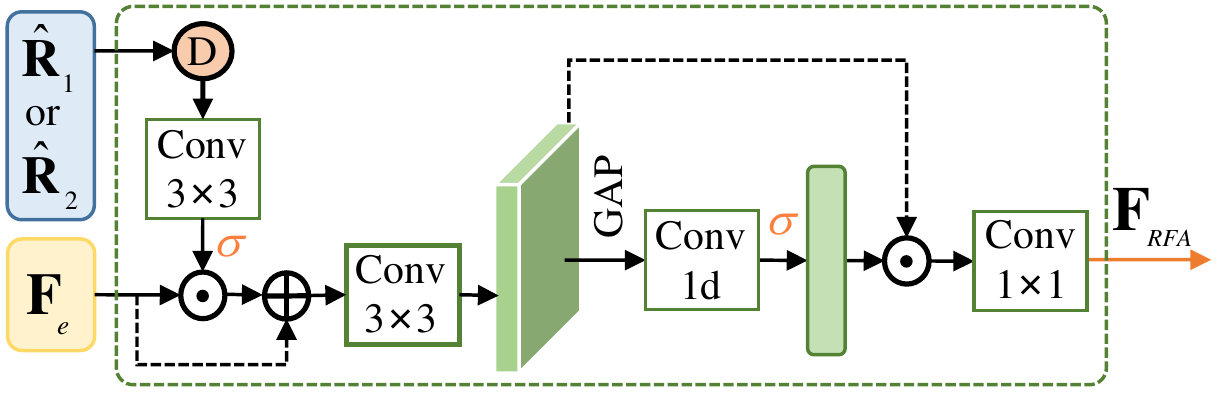}
\vspace{-2mm}
\caption{Illustration of Reconstruction-guidance Feature Aggregation (RFA) module, which can aggregate the obtained genuine compact visual patterns, such that the forgery detection capability on unknown patterns can be improved.}
\label{figure4}
\end{figure}
As shown in Figure~\ref{figure1}, we first calculate the differences between two reconstructed images (\ie, ${{\bf{\hat X}}_1}$ and ${{\bf{\hat X}}_2}$) and the original input image $\bf{X}$ in discrepancy-aggregation detector. The pixel-level difference masks are expressed as:
\begin{equation}
\label{eq9}
    {{\bf{\hat R}}_1} = \left| \bf{X}-{{\bf{\hat X}}_1} \right|, {{\bf{\hat R}}_1} = \left| \bf{X}-{{\bf{\hat X}}_2} \right|,
\end{equation}
where $\left| \cdot \right|$ refers to the absolute value function. Then for RFA in Figure~\ref{figure4}, given difference masks ${{\bf{\hat R}}_1}$ or ${{\bf{\hat R}}_2}$ and the summation of textural discrepancy information and encoding feature ${{\bf{F}}_e} = {{\bf{D}}_4} \oplus {{\bf{F}}_5}$, we perform an element-wise multiplication between them with a residual connection and a 3 $\times$ 3 convolution to obtain the fused features ${{\bf{F}}_{d}}$:
\begin{equation}
\label{eq10}
    {{\bf{F}}_{d}} = {f_{c3}}({{\bf{F}}_e} \odot (\sigma ({f_{c3}}(D({\hat R_{1/2}})))) \oplus {{\bf{F}}_e}),
\end{equation}
where $D$ denotes down-sampling and ${f_{c3}}$ is 3 $\times$ 3 convolution. $\sigma$ is a sigmoid function and $\oplus$ is element-wise addition. To enhance feature representations, inspired~\cite{wang2020eca}, we aggregate the features ${{\bf{F}}_{d}}$ using a channel-wise global average pooling (GAP). Then the channel weight is obtained by the 1D convolution followed by a sigmoid function. Finally, the channel attention is multiplied with the input features ${{\bf{F}}_{d}}$ to obtain the final output ${\bf{F}}_{RFA}$, \ie,
\begin{equation}
\label{eq11}
    {{\bf{F}}_{RFA}} = {f_{c1}}(\sigma ({f_{1d}}(GAP({{\bf{F}}_{d}}))) \odot {{\bf{F}}_{d}}),
\end{equation}
where ${f_{c1}}$ is 1 $\times$ 1 convolution and ${f_{1d}}$ is 1D convolution.
\begin{table*}[t] 
\small
\centering
\setlength{\tabcolsep}{8pt}{
\renewcommand\arraystretch{1.1}
\begin{tabular}{r|c|cccccccccc}
\toprule
\multirow{2}{*}{Methods}  & \multirow{2}{*}{Pub./Year} & \multicolumn{2}{c}{FF++ HQ} & \multicolumn{2}{c}{FF++ LQ} & \multicolumn{2}{c}{Celeb-DF} & \multicolumn{2}{c}{WLD} & \multicolumn{2}{c}{DFDC}  \\  \cmidrule(lr){3-4} \cmidrule(lr){5-6} \cmidrule(lr){7-8} \cmidrule(lr){9-10} \cmidrule(lr){11-12}  
&   & \multicolumn{1}{c}{ACC } & \multicolumn{1}{c}{AUC } & \multicolumn{1}{c}{ACC } & \multicolumn{1}{c}{AUC } & \multicolumn{1}{c}{ACC } & \multicolumn{1}{c}{AUC } & \multicolumn{1}{c}{ACC } & \multicolumn{1}{c}{AUC } & \multicolumn{1}{c}{ACC } & \multicolumn{1}{c}{AUC } \\ \midrule
Xception   & ICCV'19 & 95.73 & 96.30 & 86.86 & 89.30 & 97.90  &  99.73  & 77.25 &  86.76  & 79.35 & 89.50   \\
Two branch & ECCV'20 & 96.43 & 98.70 & 86.34 & 86.59 &  –  &   –  &  – &   –  & – & –    \\
Add-Net    & ACM MM'20   & 96.78 & 97.74 & 87.50 & 91.01 & 96.93  &  99.55  & 76.25 &  86.17  & 78.71 & 89.85     \\
F$^3$-Net  & ECCV'20 & 97.52 & 98.10 & \textcolor{blue}{90.43} & 93.30 & 95.95  &  98.93  & 80.66 &  87.53  & 76.17 & 88.39     \\
SPSL       & CVPR'21 & 91.50 & 95.32 & 81.57 & 82.82 &  –  &   –  &  – &   –  & – & –    \\
RFM        & CVPR'21 & 95.69 & 98.79 & 87.06 & 89.83 & 97.96  &  \textcolor{green}{99.94}  & 77.38 &  83.92  & \textcolor{blue}{80.83} & 89.75     \\
Freq-SCL   & CVPR'21 & 96.69 & 99.28 & 89.00 & 92.39 &  –  &   –  &  – &   –  & – & –    \\
MAT   & CVPR'21 & \textcolor{blue}{97.60} & 99.29 & 88.69 & 90.40 & 97.92  &  \textcolor{green}{99.94}  & 82.86 &  90.71  & 76.81 & \textcolor{blue}{90.32}   \\
RECCE      & CVPR'22 & 97.06 & \textcolor{blue}{99.32} & \textcolor{green}{91.03} & \textcolor{green}{95.02} & \textcolor{green}{98.59}  &  \textcolor{green}{99.94}  & \textcolor{blue}{83.25} &  \textcolor{green}{92.02}  & \textcolor{green}{81.20} & \textcolor{green}{91.33}    \\
ITA-SIA        & ECCV'22 & \textcolor{green}{97.64} & \textcolor{green}{99.35} & 90.23 & \textcolor{blue}{93.45} & \textcolor{blue}{98.48}  &  \textcolor{red}{99.96}  & \textcolor{green}{83.95} &  \textcolor{blue}{91.34}  &  – &  –    \\

\rowcolor{ours}
DisGRL & IJCAI'23 &  \textcolor{red}{97.69} &  \textcolor{red}{99.48} & \textcolor{red}{91.27} & \textcolor{red}{95.19} & \textcolor{red}{98.71}  &  \textcolor{blue}{99.91} & \textcolor{red}{84.53} &  \textcolor{red}{93.27}  & \textcolor{red}{82.35} & \textcolor{red}{92.50}    \\
\bottomrule                       
\end{tabular}}
\vspace{-2mm}
\caption{Intra-dataset evaluation and result comparisons on four benchmarks. ``HQ'' and ``LQ'' denote the High-Quality version and the Low-Quality version of the corresponding dataset, respectively.
The top three results are highlighted in \textcolor{red}{red}, \textcolor{green}{green}, and \textcolor{blue}{blue}, respectively.}
\label{table:1}
\vspace{-2mm}
\end{table*}

\subsection{Loss Function}
\label{sec3:4}
DisGRL has two kinds of supervision: the image-level binary classification label based on the cross-entropy loss (\ie, ${L_{cls}}$), and the pixel-level reconstruction learning label.
During training, we employ the reconstruction loss ($L_{r1}$ and $L_{r2}$)~\cite{Recce} between real images and their two reconstructed images. Besides, a metric-learning loss (\ie, ${L_m}$)~\cite{Recce} based on ${{\bf{F}}_5}$ is used to enhance the reconstruction difference to facilitate model learning. Thus, the total loss can be expressed as: 
\begin{equation}
\label{eq12}
    {L_{total}} = {L_{cls}} + {\lambda _1}{L_{r1}} + {\lambda _2}{L_{r2}} + {\lambda _3}{L_m},
\end{equation}
where ${\lambda}$ is a trade-off hyper-parameter for loss balance.
\section{Experiments}
\label{sec4}
\subsection{Experimental Settings}
\label{sec4-1}
\noindent\textbf{Datasets.}
To facilitate a fair result comparison with state-of-the-art methods, we conducted experiments on four fundamental yet challenging face forgery datasets, including FaceForensics++ (FF++)~\cite{rossler2019faceforensics++}, Celeb-DF~\cite{celeb-df}, WLD~\cite{wilddeepfake}, and DFDC~\cite{dfdc}. Due to the page limit, details of each dataset are given in supplementary materials. 

\noindent\textbf{Implementation Details.} 
We implemented our model on the PyTorch framework and used Xception~\cite{chollet2017xception} pre-trained on ImageNet~\cite{DengDSLL009} as our mainstream backbone.  
The input face images are resized into 299 $\times$ 299 and augmented by random horizontal flipping. 
In the training phase, the batch size is set to 32, and Adam optimizer~\cite{kingma2014adam} with learning rate 1e-4, and weight decay 1e-5 are adopted to optimize the model. The step learning rate strategy with a gamma of 0.5 is utilized to adjust the learning rate. Following~\cite{Recce}, ${\lambda_1}$, ${\lambda_2}$, and ${\lambda_3}$ in Eq. (12) are empirically set to 0.1. 

\noindent\textbf{Evaluation Metrics.}
In this work, we reported results on the commonly used three evaluation metrics~\cite{Recce,sun2022dual,zhuang2022uia,zhao2021multi}, including Accuracy (ACC), Area Under the Curve (AUC), and Equal Error Rate (EER). 

\subsection{Quantitative Results}
To demonstrate the effectiveness of our proposed method, we compare it with the state-of-the-art methods, \ie, 
Xception~\cite{rossler2019faceforensics++},
Two-branch~\cite{masi2020two},
SPSL~\cite{liu2021spatial},
RFM~\cite{wang2021representative},
Freq-SCL~\cite{li2021frequency},
Add-Net~\cite{wilddeepfake},
F${^3}$-Net~\cite{qian2020thinking},
MAT~\cite{zhao2021multi},
RECCE~\cite{Recce},
ITA-SIA~\cite{sun2022information},
Multi-task~\cite{nguyen2019multi},
MLDG~\cite{li2018learning},
LTW~\cite{sun2021domain}, and DCL~\cite{sun2022dual}. For a fair comparison, all experimental results of these methods which we employ for comparisons are either explicitly cited from works or generated by models that are retrained with open-source codes.

\noindent\textbf{Intra-Dataset Evaluation.}
Table~\ref{table:1} shows result comparisons with our DisGRL against 10 competitors under intra-dataset evaluations. We can observe that DisGRL consistently outperforms other models on FF++~\cite{rossler2019faceforensics++}, WLD~\cite{wilddeepfake}, and DFDC~\cite{dfdc}. Especially on the challenging WLD, our method still surpasses the second-best RECCE by 1.25\% in terms of AUC. This suggests that when confronted with more identities from real-world scenes, our method owns the superior ability to detect discrepancies between real faces and fake ones. 
On Celeb-DF~\cite{celeb-df}, though ITA-SIA achieves the highest AUC, our DisGRL still achieves comparable results on the other datasets, especially on the low-quality setting of the FF++ ($\uparrow$ 1.74\%). 
Different from ITA-SIA which introduces a self-information metric to enhance the feature representation, DisGRL produces a more robust representation through double-head reconstruction, which works well in conjunction with single reconstruction for forgery detection. 

\begin{table}[t] 
\small
\centering
\setlength{\tabcolsep}{4.5pt}{ 
\renewcommand\arraystretch{1.1}
\begin{tabular}{r|cccccc}
\toprule
\multirow{2}{*}{Methods}  &  \multicolumn{2}{c}{Celeb-DF} & \multicolumn{2}{c}{WLD} & \multicolumn{2}{c}{DFDC}  \\  \cmidrule(l){2-3} \cmidrule(l){4-5} \cmidrule(l){6-7}  
&  \multicolumn{1}{c}{AUC  $\uparrow$} & \multicolumn{1}{c}{EER $\downarrow$} & \multicolumn{1}{c}{AUC  $\uparrow$} & \multicolumn{1}{c}{EER $\downarrow$} & \multicolumn{1}{c}{AUC $\uparrow$} & \multicolumn{1}{c}{EER $\downarrow$}   \\ \midrule
Xception   & 61.80 & 41.73 & 62.72 & 40.65 & 63.61  & 40.58  \\
F$^3$-Net  & 61.51   & 42.03 & 57.10 & 45.12 & 64.60 & 39.84  \\
Add-Net    & 65.29 & 38.90 & 62.35 & 41.42 & 64.78 & 40.23  \\
RFM        & 65.63 & 38.54 & 57.75 & 45.45 & 66.01 & 39.05    \\
MAT   & 67.02 & 37.90 & 59.74 & 43.73 & 68.01 & 37.17  \\
RECCE      & 68.71 & 35.73 & 64.31 & 40.53 & 69.06 & 36.08  \\
\rowcolor{ours}
DisGRL      & \textbf{70.03} &  \textbf{34.23} &  \textbf{66.73} & \textbf{39.24} & \textbf{70.89} & \textbf{34.27}   \\
\bottomrule                       
\end{tabular}
\vspace{-2mm}
\caption{Cross-dataset result evaluation on FF++ (LQ), Celeb-DF, WLD, and DFDC in terms of AUC (\%) and EER (\%).}
\label{table:2}}
\vspace{-3mm}
\end{table}

\noindent\textbf{Cross-Dataset Evaluation.}
To explore the generalization of our method on unseen datasets compared with recent general face forgery detection methods, we focus on the more challenging cross-dataset evaluation. Table~\ref{table:2} reports the quantitative results by training the models on FF++ (LQ)~\cite{rossler2019faceforensics++} and testing them on Celeb-DF~\cite{celeb-df}, WLD~\cite{wilddeepfake}, and DFDC~\cite{dfdc}, accordingly. It can be concluded that our method achieves a certain improvement in generalization ability by taking good advantage of double-head reconstruction structures. 
In particular, the AUC score of our method on Celeb-DF ($\uparrow$ 1.32\%), WLD ($\uparrow$ 2.42\%), and DFDC ($\uparrow$ 1.83\%) datasets is enhanced when compared with RECCE. 
Overall, our method promotes the extraction of genuine compact visual patterns and can be generalized to unseen forgeries rather than modeling the pattern of the single forgery techniques.

\begin{figure*}[th]
\centering
\includegraphics[width=.95\linewidth]{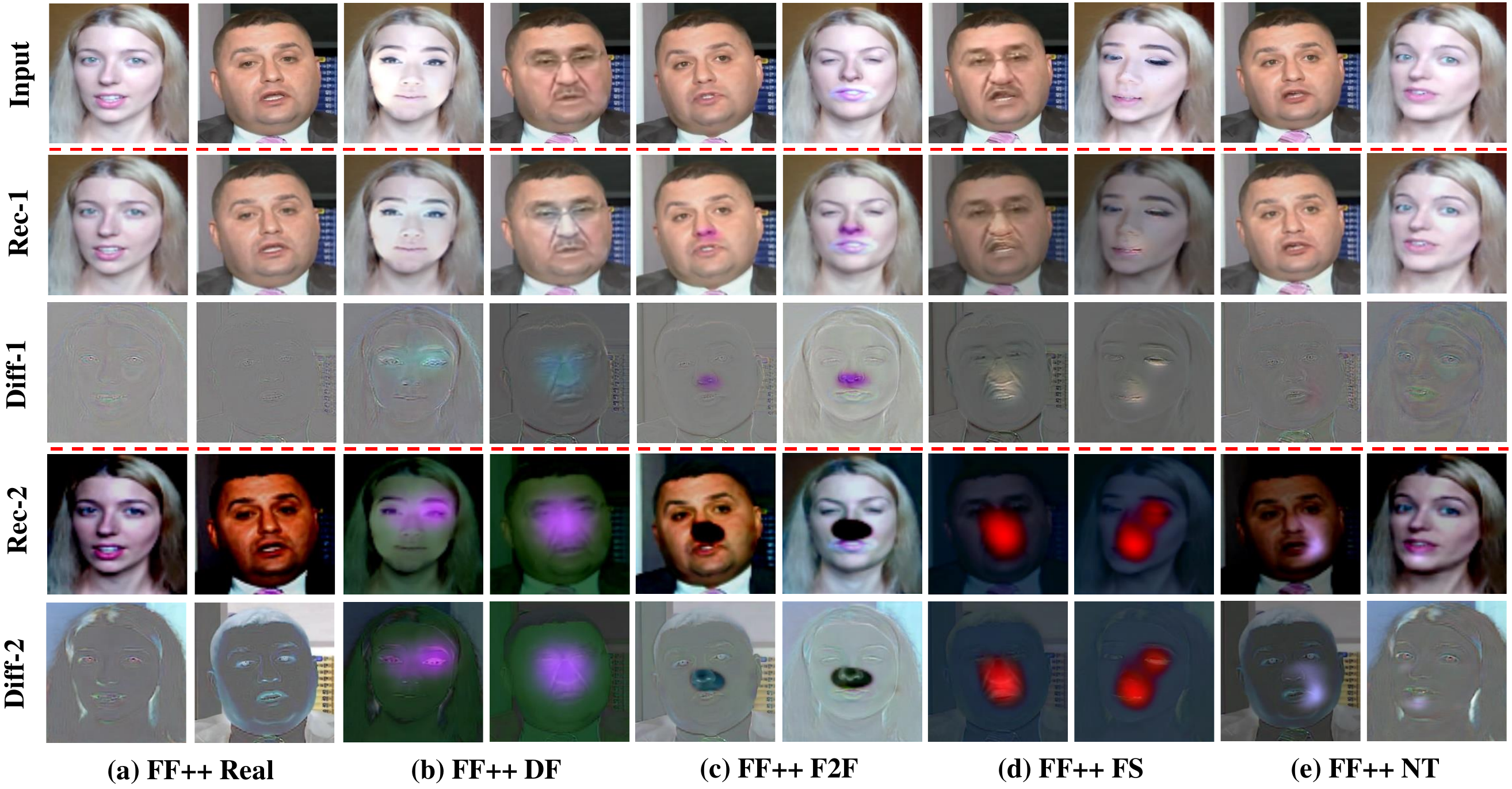}
\vspace{-3mm}
\caption{Reconstruction and differential visualization of the proposed model on the FaceForensics++ dataset. ``Rec-1'' and ``Rec-2'' are the double reconstructions results. ``Diff-1'' and ``Diff-2'' denote the corresponding pixel-level difference, respectively.}
\label{fig: 5}
\vspace{-3mm}
\end{figure*}

\noindent\textbf{Cross-Manipulation Evaluation.}
To further demonstrate the generalization among different manipulated manners, we conduct the fine-grained cross-manipulation evaluation by training a model on one specific method and testing it on all four methods listed in FF++ (LQ). As shown in Table~\ref{table:3}, our DisGRL generally outperforms the competitors in most cases, including both intra-manipulation (diagonal of the table) results and cross-manipulation. Specifically, when training on NT and testing on F2F, though MAT is equipped with EfficientNet-b4, our DisGRL based on Xception still outperforms it by a margin of 2.69\%. Additionally, a 2.41\% performance gain in terms of AUC is achieved by our method compared with RECCE, which illustrates that it is feasible to explore common features of real faces to distinguish real and fake faces. With help of the double-head reconstruction strategy and carefully designed cascaded discrepancy external attention, our method exceeds all other methods in terms of the average AUC of cross-manipulation evaluations.

\noindent\textbf{Multi-Source Manipulation Evaluation.}
Multi-source manipulation evaluation refers to situations in which the forged techniques utilized for training are not restricted to just one way. Following the LTW~\cite{sun2021domain} and DCL~\cite{sun2022dual}, we conduct experiments on the low-quality (LQ) version of
FF++~\cite{rossler2019faceforensics++} to demonstrate the practicality of our method in real-world scenarios. As shown in Table~\ref{table:4}, we can observe that our DisGRL obtains cutting-edge performance in terms of AUC and ACC on all protocols. In particular, DisGRL outperforms the recent DCL by around 7\% in the setting of GID-F2F, proving its durability and ability to ensure generalization under various scenarios.
\begin{table}[t] 
\small
\setlength{\tabcolsep}{5.5pt}{ 
\renewcommand\arraystretch{1.1}
\begin{tabular}{r|cccccc}
\toprule
Methods & Train & DF   &  F2F    &  FS   &  NT   &  CAvg.
\\ \midrule
Freq-SCL &  \multirow{4}{*}{DF} & \cellcolor[gray]{.95}98.91 & 58.90 &66.87 &63.61& 63.13 \\
MAT &   & \cellcolor[gray]{.95}99.51 & 66.41 & 67.33 & 66.01 & 66.58 \\
RECCE &   & \cellcolor[gray]{.95}99.65 & 70.66 & 74.29 & 67.34 & 70.76 \\
DisGRL &   & \cellcolor[gray]{.95}{\textbf{99.67}}   & \textbf{71.76}   &  \textbf{75.21} & \textbf{68.74}  &  \textbf{71.90}  \\  \midrule

Freq-SCL &  \multirow{4}{*}{F2F} & 67.55 &\cellcolor[gray]{.95}93.06 & 55.35 & 66.66 & 63.19 \\
MAT &   & 73.04 &\cellcolor[gray]{.95}97.96 & 65.10 & 71.88 & 70.01 \\
RECCE &      & \textbf{75.99} &\cellcolor[gray]{.95}98.06 & 64.53 & 72.32 & 70.95 \\
DisGRL &         &  75.73   &\cellcolor[gray]{.95}\textbf{98.69}       &  \textbf{65.71}    &   \textbf{74.15}    &  \textbf{71.86}     \\  \midrule

Freq-SCL &  \multirow{4}{*}{FS} & 75.90 &54.64 &\cellcolor[gray]{.95}98.37 &49.72& 60.09 \\
MAT &   & 82.33 &61.65 &\cellcolor[gray]{.95}98.82 &54.79& 66.26 \\
RECCE &   & 82.39 &64.44 &\cellcolor[gray]{.95}98.82 &56.70& 67.84 \\
DisGRL &   & \textbf{82.73} & \textbf{64.85}  &\cellcolor[gray]{.95}{\textbf{99.01}} & \textbf{56.96} & \textbf{68.18}  \\  \midrule

Freq-SCL &  \multirow{4}{*}{NT} & 79.09 & 74.21 & 53.99 &\cellcolor[gray]{.95}88.54 & 69.10 \\
MAT &   & 74.56 & 80.61 & 60.90 &\cellcolor[gray]{.95}93.34 & 72.02 \\
RECCE &   & 78.83 & 80.89 & 63.70 &\cellcolor[gray]{.95}93.63 & 74.47 \\
DisGRL &   & \textbf{80.29} & \textbf{83.30} & \textbf{65.23}  &\cellcolor[gray]{.95}{\textbf{94.10}} & \textbf{76.27} \\  
\bottomrule                       
\end{tabular}
\vspace{-2mm}
\caption{Cross-manipulation evaluation in terms of AUC (\%). Diagonal results indicate intra-domain performance. DeepFakes (DF), Face2Face (F2F), FaceSwap (FS), and NeuralTextures (NT) are four image manipulation approaches in FF++~\protect\cite{rossler2019faceforensics++}. ``CAvg.'' denotes the average of cross-manipulation evaluations.}
\label{table:3}}
\vspace{-2mm}
\end{table}

\begin{table}[t] 
\footnotesize
\setlength{\tabcolsep}{3.2pt}{ 
\begin{tabular}{r|cccccccc}
\toprule
\multirow{2}{*}{Methods}  & \multicolumn{2}{c}{GID-DF} & \multicolumn{2}{c}{GID-F2F} & \multicolumn{2}{c}{GID-FS} & \multicolumn{2}{c}{GID-NT}  \\  \cmidrule(l){2-3} \cmidrule(l){4-5} \cmidrule(l){6-7} \cmidrule(l){8-9}
&    \multicolumn{1}{c}{ACC } & \multicolumn{1}{c}{AUC } & \multicolumn{1}{c}{ACC } & \multicolumn{1}{c}{AUC } & \multicolumn{1}{c}{ACC } & \multicolumn{1}{c}{AUC } & \multicolumn{1}{c}{ACC } & \multicolumn{1}{c}{AUC } \\ \midrule
Multi-task   &  66.8 & – & 56.5 & – & 51.7  &  –  & 56.0 &  –    \\
MLDG         &  67.2 & 73.1 & 58.1 & 61.7  & 58.1   & 61.7  & 56.9 &  60.7     \\
LTW          &  69.1 & 75.6 & 65.7 & 72.4 & 62.5  &  68.1  & 58.5 &  60.8     \\
DCL          &  75.9 & 83.8 & 67.9 & 75.1 & –  &  –  & – &  –     \\
\rowcolor{ours}
DisGRL         & \textbf{77.3}       & \textbf{86.1}      & \textbf{75.8}      &    \textbf{84.3}   &   \textbf{76.9}     &   \textbf{86.3}      &  \textbf{66.3}  & \textbf{72.8}      \\
\bottomrule                       
\end{tabular}}
\vspace{-2mm}
\caption{Multi-source manipulation evaluation in terms of ACC (\%) and AUC (\%). The protocols and results are from LTW~\protect\cite{sun2021domain} and DCL~\protect\cite{sun2022dual}.}
\label{table:4}
\vspace{-2mm}
\end{table}


\subsection{Ablation Study}
To validate the effectiveness of each component, we designed several ablation experiments on the WildDeepfake dataset in varied configurations with the components added progressively. As shown in Table~\ref{table:5}, the setup model variants are as follows: for the baseline model of a), we follow the classic image classification pipeline, \ie, Xception~\cite{chollet2017xception}. b) and c) the encoder-decoder backbone with the introduction of a single-head reconstruction ({Rec}-1 or {Rec}-2) learning scheme. d), the encoder-decoder backbone equipped with double-head reconstruction ({Rec}-1 and {Rec}-2). For the model of e), we remove the RFA and adopt the element-wise addition to replace it, f) is the proposed DisGRL without DEA, and g) is our DisGRL.

\noindent\textbf{Effectiveness of DouHR.}
We can observe in Table~\ref{table:5} that the double-head reconstruction learning module performs better than the baseline model of a) and its variant b), c)~Baseline + single-head. Particularly, the model of d)~Baseline + double-head improves the performance above models of a), b), and c) by 5.71\%, 2.08\%, 1.47\% in ACC, and 4.77\%, 2.51\%, 2.04\% in AUC, respectively. Therefore, {Rec}-2, as the complementary information of {Rec}-1, aims to capture the genuine compact visual pattern of real regions and fake regions, which is beneficial to boost detection performance.

\noindent\textbf{Effectiveness of DisGE.} Then the model of e)~DisGRL w/o RFA achieves better overall performance compared with the model of d)~Baseline + double-head, especially in terms of AUC with 1.37 \% performance gains. It verifies that DEA enhances the model's efficiency to mine the forgery-sensitive visual pattern within the instance by cascading shallow and deep features in the encoder to concentrate more on image forgery cues rather than on semantic image content. Therefore, it is feasible to improve the classification learning capabilities of the detector when combined with the integrated representation collected by the decoder, leading to a larger performance increase for variation e)~DisGRL w/o RFA.

\noindent\textbf{Effectiveness of DisAD.}
The comparison between variants d)~Baseline + double-head and f)~DisGRL w/o DEA in Table~\ref{table:5} can demonstrate the effectiveness of our proposed RFA, which aggregates the obtained genuine compact visual pattern and emphasizes the probably forged regions. And combining all the proposed components can achieve the best performance in terms of ACC and AUC scores.
\begin{table}[t] 
\centering
\footnotesize
\setlength{\tabcolsep}{4.5pt}{ 
\renewcommand\arraystretch{1.1} 
\begin{tabular}{c|ccccc|cc}
\toprule
NO & Baseline  & {Rec}-1 & {Rec}-2 & DEA & RFA & ACC & AUC \\ 
 \midrule
a) & \checkmark &   &  &  &  & 77.13 & 86.21    \\
b) & \checkmark  &  \checkmark  &  &  &   & 80.76  & 88.47  \\
c) & \checkmark  &    &\checkmark  &  &   & 81.37  & 88.94  \\
d) &\checkmark  &  \checkmark  &  \checkmark &  &  & 82.84 & 90.98\\
e) &\checkmark   &  \checkmark  &  \checkmark & \checkmark &  & 83.68 & 92.35   \\
f) &\checkmark   &  \checkmark  &  \checkmark &  & \checkmark & 83.24 & 91.68  \\
g) & \checkmark   &    \checkmark    &   \checkmark     &    \checkmark &     \checkmark   &   \textbf{84.53}     & \textbf{93.27}   \\
\bottomrule                       
\end{tabular}}
\vspace{-2mm}
\caption{Ablation studies on WildDeepfake~\protect\cite{wilddeepfake} in terms of ACC (\%) and AUC (\%). ``RFA'' and ``DEA'' are the Reconstruction-guidance Feature Aggregation module and Discrepancy External Attention (DEA) block, respectively. }
\label{table:5}
\end{table}

\subsection{Visualizations}
Our proposed reconstruction learning aims to preserve more variations by building a double-head reconstruction scheme. To validate its effectiveness, as illustrated in Figure~\ref{fig: 5}, we visualize the outputs of the two reconstructions and the corresponding difference masks between the original input and reconstruction maps. We can observe that the real faces can be well reconstructed with little blurring, while the forged portions of the fake ones cannot be recovered. Difference masks, indicating possible traces of forged areas, further amplify the differences between real and forged faces. Compared with Diff-1, Diff-2 is able to additionally enhance and complement the forged areas in faces. 
For instance, NT operates around the mouth region and the response in Diff-1 of the corresponding sample is weak around the mouth region, while the value is larger in Diff-2, illustrating the importance and usefulness of an additional head for reconstruction in image forgery detection.

\section{Conclusion}
In this work, we proposed a novel image forgery detection paradigm, termed DisGRL, to improve the model learning capacity on forgery-sensitive and genuine compact visual patterns. DisGRL mainly consisted of a discrepancy-guided encoder, a decoder, a double-head reconstruction module, and a discrepancy-aggregation detector head network for image forgery classification. The advantage of DisGRL was that it can not only encode general semantic features but also enhance the forgery cues of the given image. Experimental results on four widely used face forgery datasets validated the effectiveness of our proposed method against state-of-the-art competitors on both seen and unseen forgeries. DisGRL is a general paradigm, which can be used in general image forgery detection tasks. Therefore, in the future, we will explore how to apply DisGRL to more challenging natural scene datasets in terms of quantity and quality. Besides, exploring how to use DisGRL in the forgery detection of video data is also a promising research direction. 


\newpage
\section*{Supplementary Materials}
This supplementary includes an introduction of the Metric-Learning Loss, detailed introductions of our utilized datasets, quantitative results with multi-source manipulation evaluation, analysis of classification decision, computational complexity and time cost, and analysis of robustness.
\subsection*{\textbf{\emph{S1.}}~Metric-Learning Loss}
\label{S1}
{\color{red}{This supplementary is for Sec. 3.4 ``Loss Function'' of the main paper.}} 
During the reconstruction process, in addition to the reconstruction difference, we adopt a metric-learning loss (\ie, ${L_m}$)~\cite{Recce} based on encoder output to make real images close while real and fake images far away in the embedding space.
Let ${{\bf{F}}_5 \in {\mathbb{R}^{\textit{C}_5 \times \textit{H}_5 \times \textit{W}_5}}}$ denote the output of the last encoder block. We apply the global average pooling operation to ${{\bf{F}}_5}$ and obtain the feature vector ${{\bf{\bar F}}_5} \in {\mathbb{R}^{\textit{C}_5}}$ for each input sample. 
The metric-learning loss is:
\begin{equation}
\begin{split}
{L_m} = & \frac{1}{{{N_{RR}}}}\sum\limits_{i \in R,j \in R} {d({{\bf{\bar F}}_5^i},{{\bf{\bar F}}_5^j})}  - \frac{1}{{{N_{RF}}}}\sum\limits_{i \in R,j \in F} {d({{\bf{\bar F}}_5^i},{{\bf{\bar F}}_5^j})},
\end{split}
\label{eqS1}
\end{equation}
where $R$ and $F$ indicate the set of real and fake samples. ${N_{RR}}$ and ${N_{RF}}$ are the total number of (real, real) sample pairs and (real, fake) sample pairs, respectively. The first part in ${L_m}$ encourages learning compact representations from genuine faces while the second part emphasizes the differences between real and fake samples. $d( \cdot , \cdot )$ denote a pair-wise distant function based on the cosine distance:
\begin{equation}
\label{eqS2}
d({\bf{a}},{\bf{b}}) = \frac{{1 - \frac{{\bf{a}}}{{{{\left\| {\bf{a}} \right\|}_2}}} \cdot \frac{{\bf{b}}}{{{{\left\| {\bf{b}} \right\|}_2}}}}}{2}.
\end{equation}

\begin{figure}[t]
\centering
\includegraphics[width=1\linewidth]{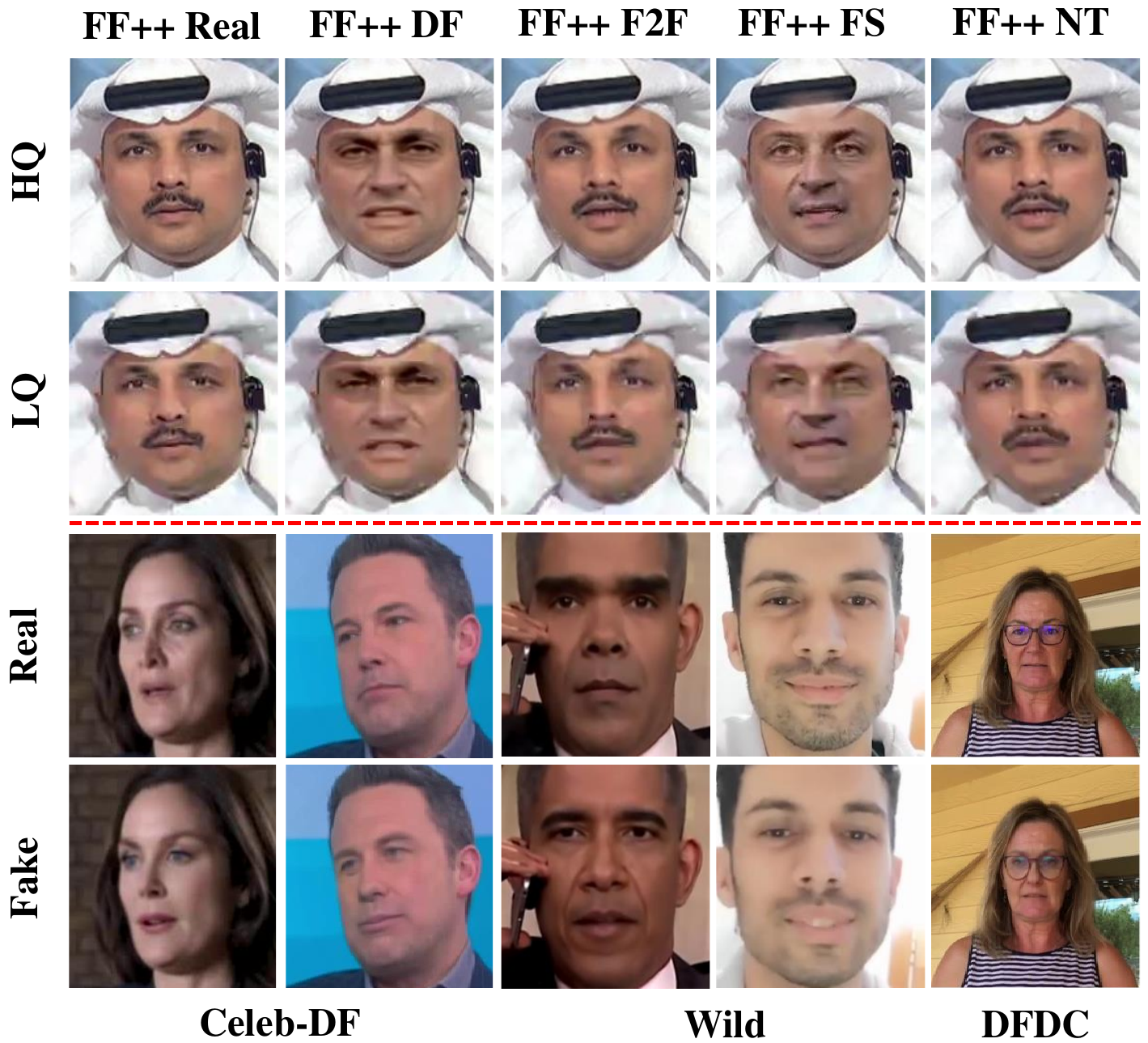}
\caption{Some examples from FF++~\protect\cite{rossler2019faceforensics++}, Celeb-DF~\protect\cite{celeb-df}, WLD~\protect\cite{wilddeepfake}, and DFDC~\protect\cite{dfdc}. ``HQ'' and ``LQ'' denote the High-Quality version and the Low-Quality version of the FF++~\protect\cite{rossler2019faceforensics++} dataset, respectively. ``DF'', ``F2F'', ``FS'', and ``NT'' represent DeepFakes, Face2Face, FaceSwap, and NeuralTextures forgery techniques in FF++~\protect\cite{rossler2019faceforensics++} dataset.}
\label{fig6}
\end{figure}

\subsection*{\textbf{\emph{S2.}}~Dataset Details}
\label{S2}
{\color{red}{This supplementary is for Sec. 4.1 ``Experimental Settings'' of the main paper.}} As shown in Figure~\ref{fig6}, we evaluate our proposed method and existing approaches on FF++~\cite{rossler2019faceforensics++}, Celeb-DF~\cite{celeb-df}, WLD~\cite{wilddeepfake}, and DFDC~\cite{dfdc}. 
\textbf{\romannumeral1)~FF++} is the most widely used dataset consisting of 1,000 real videos and corresponding 4,000 fake videos which are generated by four facial image manipulation methods, namely DeepFakes (DF), Face2Face (F2F), FaceSwap (FS), and NeuralTextures (NT). It provides two kinds of video quality: High-Quality denoted by HQ (quantization of 23) and Low-Quality denoted by LQ (quantization of 40). \textbf{\romannumeral2)~Celeb-DF}~\cite{celeb-df} is another widely-used dataset, which contains 590 real videos and 5,639 fake videos. \textbf{\romannumeral3)~WDF}~\cite{wilddeepfake} includes 3,805 real face sequences and 3,509 fake face sequences, which has a diversified distribution of real-world scenarios. \textbf{\romannumeral4)~DFDC}~\cite{dfdc} is by far the largest currently and publicly-available face swap video dataset which totally contains 128,154 facial videos of 960 agreeing subjects. We follow the previous works~\cite{Recce,zhao2021multi}, which use RetinaFace to extract faces for all real/fake video frames in FaceForensics++~\cite{rossler2019faceforensics++}, Celeb-DF~\cite{celeb-df}, and DFDC~\cite{dfdc} datasets.

\begin{table}[ht] 
\footnotesize
\setlength{\tabcolsep}{3.2pt}{ 
\begin{tabular}{r|cccccccc}
\toprule
\multirow{2}{*}{Methods}  & \multicolumn{2}{c}{GID-DF} & \multicolumn{2}{c}{GID-F2F} & \multicolumn{2}{c}{GID-FS} & \multicolumn{2}{c}{GID-NT}  \\  \cmidrule(l){2-3} \cmidrule(l){4-5} \cmidrule(l){6-7} \cmidrule(l){8-9}
&    \multicolumn{1}{c}{ACC } & \multicolumn{1}{c}{AUC } & \multicolumn{1}{c}{ACC } & \multicolumn{1}{c}{AUC } & \multicolumn{1}{c}{ACC } & \multicolumn{1}{c}{AUC } & \multicolumn{1}{c}{ACC } & \multicolumn{1}{c}{AUC } \\ \midrule
Multi-task   &  70.3 & – & 58.7 & – & 49.7  &  –  & 60.3 &  –    \\
MLDG         &  84.2 & 91.8 & 63.4 & 77.1  & 52.7   & 60.9  & 62.1 &  78.0     \\
LTW          &  85.6 & 92.7 & 65.6 & 80.2 & 54.9  &  64.0  & 65.3 &  77.3     \\
DCL          &  87.7 & 94.9 & 68.4 & 82.9 & –  &  –  & – &  –     \\
\rowcolor{ours}
DisGRL         & \textbf{88.3}       & \textbf{95.7}      & \textbf{74.7}      &    \textbf{89.4}   &   \textbf{58.4}     &   \textbf{67.9}      &  \textbf{67.6}  & \textbf{84.6}      \\
\bottomrule                       
\end{tabular}}
\caption{Multi-source manipulation evaluation in terms of ACC (\%) and AUC (\%). The protocols and results are from LTW~\protect\cite{sun2021domain} and DCL~\protect\cite{sun2022dual}. GID-DF indicates training on the remaining three facial manipulation techniques of FF++ (HQ) and testing on DeepFakes class. The same for the other setting.}
\label{table:6}
\end{table}
\subsection*{\textbf{\emph{S3.}}~Quantitative Results with Multi-Source Manipulation Evaluation}
\label{S3}
{\color{red}{This supplementary is for Sec. 4.2 ``Quantitative Results'' of the main paper.}} In Sec. 4.2, we conduct multi-source manipulation evaluation on FF++ (LQ) in terms of ACC and AUC. In this part, we provided more quantitative results on multi-source manipulation evaluation to better demonstrate the practicality of our method. Specifically, our method is compared to the baselines on the GID benchmarks with images of high quality (HQ). As shown in Table~\ref{table:6}, we can observe that our DisGRL obtains state-of-the-art performance on all protocols in terms of AUC and ACC. In particular, our method outperforms the recent DCT by around 6\% on the GID-F2F of FF++ (HQ) in terms of AUC and ACC scores, which shows our method can guarantee generalization under different conditions and further demonstrates the robustness of our framework.

\begin{figure}[t]
\centering
\includegraphics[width=1\linewidth]{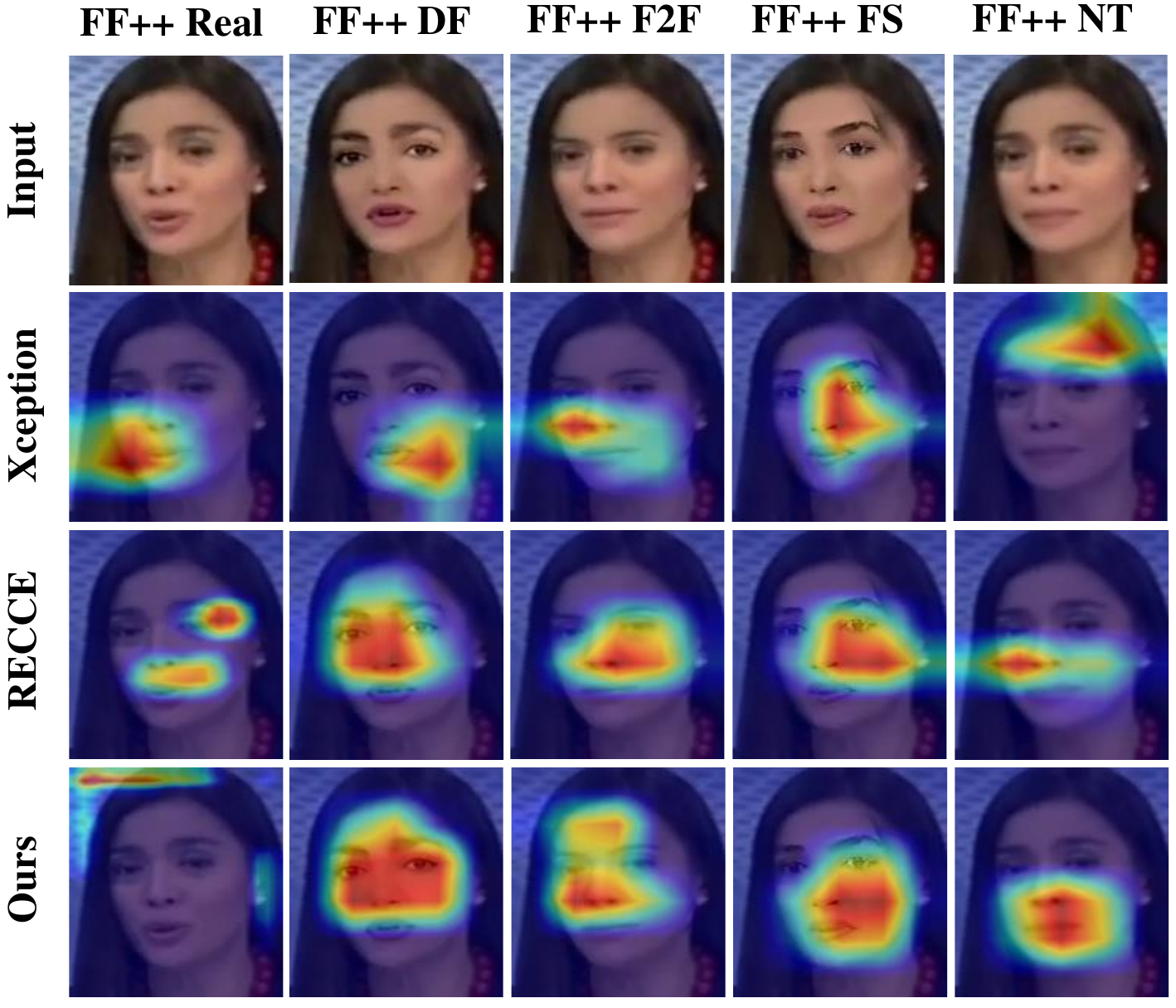}
\caption{The Grad-CAM~\protect\cite{selvaraju2017grad} visualization.}
\label{fig7}
\end{figure}
\begin{figure*}[t]
\centering
\includegraphics[width=1\linewidth]{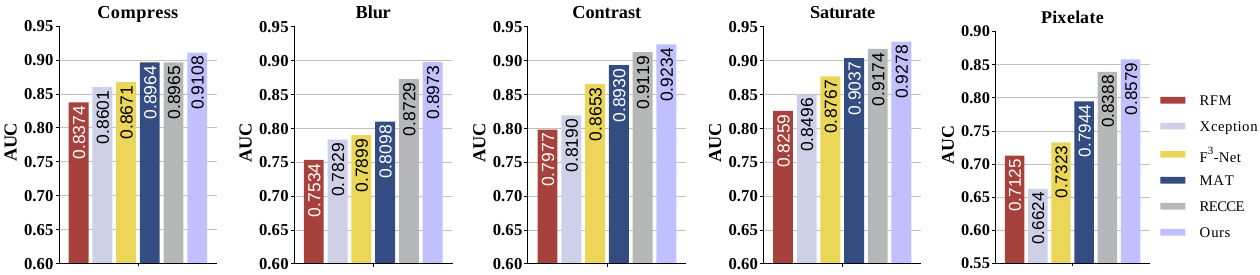}
\caption{Robustness evaluation on WildDeepfake dataset. AUC (\%) scores are reported.}
\label{fig8}
\end{figure*}
\subsection*{\textbf{\emph{S4.}}~Analysis of Classification Decision}
\label{S4}
{\color{red}{This supplementary is for Sec. 4 ``Experiments'' of the main paper.}}
To better explore which region our method focuses on that affects the model classification decision, we provide the Grad-CAM~\cite{selvaraju2017grad} visualization on FF++ in Figure~\ref{fig7}. The warm color indicates the regions that respond strongly to the prediction of forgery. It is evident that the baseline method Xception focuses mainly on partial face regions in the image for classification without considering the realness of the face. In contrast, RECCE and our method are significantly better at producing distinguishable heatmaps for real and fake faces. Our method, however, not only shows little response to the real face but also produces a more thorough reaction to various manipulation techniques. For instance, both for DF and NT, our heatmaps concentrate on the main facial region and the mouth region with a larger response area, which facilitates the model to make the appropriate classification decision. This is consistent with the performance of our DisGRL in quantitative evaluation and reveals the superior classification ability of our method for different forgeries.
 
\subsection*{\textbf{\emph{S5.}}~Computational Complexity and Time Cost}
\label{S5}
{\color{red}{This supplementary is for Sec. 4 ``Experiments'' of the main paper.}}
We compared the computational complexity and time cost of our technique to existing methods. We would like to emphasize that metrics such as \textbf{\emph{average training time, as well as inference time per input image, are better indicators than model parameters and FLOPs}}. This is because many approaches adopt complex tricks that cannot be accurately measured with only parameters and FLOPs. As shown in Table~\ref{table:7}, our DisGRL consumes less time for training and inference compared to RECCE~\cite{Recce}, LTW~\cite{sun2021domain}, and MVSS-Net~\cite{chen2021image}, despite having more parameters and FLOPs than RECCE. Specifically, DisGRL(w/o Rec-2) takes only approximately 8.4 minutes for training one epoch and 15.4ms per image for evaluation, while DisGRL(w/o Rec-1) takes 9.2 minutes and 16.5ms, respectively. However, we acknowledge that the design of SAM and AFS requires slightly more parameters and FLOPs. Therefore, we plan to modify our approach in future work to optimize the computational complexity and achieve a better balance between accuracy and complexity. In summary, we believe that the efficiency of DisGRL is acceptable and practically usable.

\begin{table}[h!] 
\small
\centering
\footnotesize
\setlength{\tabcolsep}{0.9pt}{ 
\begin{tabular}{l|cccc}
\toprule
Methods & Inference time  & Training time & \textcolor{gray}{Params}  & \textcolor{gray}{FLOPs}   \\ 
 \midrule
RECCE    & 29.7ms & 33.5mins  & \textcolor{gray}{30.07M} & \textcolor{gray}{15.11G}   \\
LTW    & 23.5ms & 17.9mins  & \textcolor{gray}{98.78M} & \textcolor{gray}{63.25G}   \\
MVSS-Net  & 21.9ms & 19.4mins  & \textcolor{gray}{146.81M} & \textcolor{gray}{67.94G}   \\
\textbf{DisGRL(w/o Rec-2)}  & 15.4ms & 8.4mins & \textcolor{gray}{89.72M}  & \textcolor{gray}{54.86G}\\
\textbf{DisGRL(w/o Rec-1)}   & 16.5ms & 9.2mins & \textcolor{gray}{82.34M}  & \textcolor{gray}{48.05G} \\
\textbf{DisGRL}   & 20.7ms & 10.1mins  & \textcolor{gray}{95.87M}  &  \textcolor{gray}{58.09G}\\
\bottomrule                       
\end{tabular}}
\caption{Results are obtained on a single NVIDIA GeForce RTX 3090 GPU. The training time is the average over 10 epochs, and the inference time is the average of over 14,000 test images from FF++ (Neural Textures) dataset with image size 299 $\times$ 299. ``M'' denotes million, ``ms'' denotes milliseconds, and ``mins'' denotes minutes. }
\label{table:7}
\end{table}

\subsection*{\textbf{\emph{S6.}}~Analysis of Robustness}
\label{S6}
{\color{red}{This supplementary is for Sec. 4 ``Experiments'' of the main paper.}}
 Figure~\ref{fig8} shows the robustness comparisons of different face forgery detection methods against several typical perturbations suggested by~\cite{jiang2020deeperforensics,Recce}, \ie~image compression, Gaussian blur, contrast jitter, saturate jitter, and pixelation. These results intuitively show that our DisGRL is less affected by image compression, contrast jitter and saturate jitter distortion operations and exhibits more robust performance, while it is more sensitive to Gaussian blur and pixelation distortion operations. This may be because we used the discrepancy-guided module to dig into the minute forgery artifacts and compression errors, as well as to gather additional perceptual clues containing the local and global details of the tampered region. Moreover, the robustness of the DisGRL is manifested by the fact that it continuously outperforms previous methods against multiple distortion attacks by a large margin, \ie, 2.44\% for Gaussian blur and 1.91\% for pixelation over the state-of-the-art RECCE~\cite{Recce}.

\end{document}